\def\year{2017}\relax
\documentclass[letterpaper]{article}
\usepackage{aaai17}
\usepackage{times}
\usepackage{helvet}
\usepackage{courier}
\usepackage{amsfonts,amssymb}
\usepackage{amsmath}
\usepackage[utf8]{inputenc} 
\usepackage[T1]{fontenc} 
\usepackage{url} 
\usepackage{booktabs} 
\usepackage{nicefrac} 
\usepackage{microtype} 
\usepackage{algorithm}
\usepackage{algorithmic}
\usepackage{graphicx}

\usepackage{times}
\usepackage{subfigure}
\usepackage[toc,page]{appendix}

\newtheorem{theorem}{Theorem}[section]

\newtheorem{corollary}[theorem]{Corollary}
\begin{document}
%
	\title{Asynchronous Stochastic Proximal Optimization Algorithms with Variance Reduction}
	\author{Qi Meng\textsuperscript{1}, Wei Chen\textsuperscript{2}, Jingcheng Yu\textsuperscript{3}, Taifeng Wang\textsuperscript{2}, Zhi-Ming Ma\textsuperscript{4}, Tie-Yan Liu\textsuperscript{2}\\
		\textsuperscript{1}
		School of Mathematical Sciences, Peking University,
		qimeng13@pku.edu.cn\\\textsuperscript{2}Microsoft Research, $\{$wche, taifengw, tie-yan.liu$\}$@microsoft.com\\\textsuperscript{3}Fudan University, JingchengYu.94@gmail.com\\\textsuperscript{4}Academy of Mathematics and Systems Science, Chinese Academy of Sciences, mazm@amt.ac.cn
	}
\maketitle
\begin{abstract}
	Regularized empirical risk minimization (R-ERM) is an important branch of machine learning, since it constrains the capacity of the hypothesis space and guarantees the generalization ability of the learning algorithm. Two classic proximal optimization algorithms, i.e., proximal stochastic gradient descent (ProxSGD) and proximal stochastic coordinate descent (ProxSCD) have been widely used to solve the R-ERM problem. Recently, variance reduction technique was proposed to improve ProxSGD and ProxSCD, and the corresponding ProxSVRG and ProxSVRCD have better convergence rate. These proximal algorithms with variance reduction technique have also achieved great success in applications at small and moderate scales. However, in order to solve large-scale R-ERM problems and make more practical impacts, the parallel version of these algorithms are sorely needed. In this paper, we propose asynchronous ProxSVRG (Async-ProxSVRG) and asynchronous ProxSVRCD (Async-ProxSVRCD) algorithms, and prove that Async-ProxSVRG can achieve near linear speedup when the training data is sparse, while Async-ProxSVRCD can achieve near linear speedup regardless of the sparse condition, as long as the number of block partitions are appropriately set. We have conducted experiments on a regularized logistic regression task.  The results verified our theoretical findings and demonstrated the practical efficiency of the asynchronous stochastic proximal algorithms with variance reduction.
\end{abstract}

\section{Introduction}

In this paper, we focus on the regularized empirical risk minimization (R-ERM) problem, whose objective is a finite sum of smooth convex loss functions $f_i(x)$ plus a non-smooth regularization term $R(x)$, i.e.,
{\small\begin{equation}\label{eq1}
	\min_{x\in \mathbb{R}^d}P(x)= F(x)+R(x)=\frac{1}{n}\sum_{i=1}^{n}f_i(x)+R(x).
	\end{equation}}
In particular, in the context of machine learning, $f_i(x)$ and $R(x)$ are defined as follows. Suppose we are given a collection of training data $(a_1,b_1)$,...,$(a_n,b_n)$, where each $a_i\in \mathbb{R}^d$ is an input feature vector and $b_i\in \mathbb{R}$ is the output variable. The loss function $f_i(x)$ measures the fitness of the model $x$ on training data $(a_i, b_i)$. Different learning tasks may use different loss functions, such as the least square loss $ \frac{1}{2}(a_i^Tx-b_i)^2$ for regression and the logistic loss  $\log(1+\exp(-b_ia_i^Tx))$ for classification. The regularization term is used to constrain the capacity of the hypothesis space. For example, the non-smooth $L_1$ regularization term is widely used.

In order to solve the R-ERM problem, the proximal stochastic gradient descent method (ProxSGD) has been widely used, which exploits the additive nature of the empirical risk function and updates the model based on the gradient which is calculated at randomly sampled training data. However, the random sampling in ProxSGD introduces non-negligible variance, which makes that we need to use a decreasing step size (also known as learning rate) to guarantee the algorithm's convergence, and the convergence rate is only sublinear  \cite{langford2009sparse,rakhlin2011making}. To tackle this problem, people have developed a set of new technologies. For example, in \cite{xiao2014proximal}, a variance reduction technique was introduced to improve ProxSGD and a new algorithm called ProxSVRG was proposed. It has been proven that even with a constant step size, ProxSVRG can achieve linear convergence rate.

Proximal stochastic coordinate descent (ProxSCD) is another method which is used to solve the R-ERM problem \cite{shalev2011stochastic}. Since the variance introduced by the coordinate sampling asymptotically goes to zero, the ProxSCD attains linear convergence rate when the objective function $P(x)$ is strongly convex \cite{wright2015coordinate}. However, ProxSCD still requires that all component functions in the empirical risk are accessible in each iteration, which is time consuming. In \cite{zhao2014accelerated}, a new algorithm called ProxSVRCD (also known as MRBCD) was proposed to improve ProxSCD. This algorithm, in addition to randomly samples a block of coordinates, also randomly samples training data in each iteration and uses the variance reduction technique. It has been proven that ProxSVRCD can achieve linear convergence rate and outperform ProxSCD by a lower iteration complexity.

While the aforementioned new algorithms (i.e., ProxSVRG and ProxSVRCD) have both good theoretical properties and empirical performances, the investigations on them were mainly conducted in the sequential (single-machine) setting. In this big data era, we usually need to deal with very large scale R-ERM problems. In this case, sequential algorithms usually cost too much time. To tackle the challenge, parallelization of these algorithms are sorely needed. Recently literature research in parallel method tend to use asynchronous parallelization due to its high efficient in system \cite{dean2012large,recht2011hogwild}. We are interested in asynchronous parallel implementations of the aforementioned stochastic proximal algorithms with variance reduction, which are, however, not well studied in the literature, to the best of our knowledge.

For asynchronous ProxSVRG (Async-ProxSVRG), we consider the \emph{consistent read} setting, in which we ensure the atomic pull and push of the whole parameter for the local workers. For asynchronous ProxSVRCD (Async-ProxSVRCD), since the updates are performed over coordinate blocks, we only ensure the atomic pull and push of a coordinate block of the parameter for local workers for the sake of system efficiency. Comparing with Async-ProxSVRG setting, we name it as \emph{inconsistent read} setting. We conduct theoretical analysis for Async-ProxSVRG and Async-ProxSVRCD. According to our results: (1) Async-ProxSVRG can achieve near linear speedup with respect to the number of local workers, when the input feature vectors are sparse; (2) If the data are non-sparse, ProxSVRCD can still achieve near linear speedup, when the block size is small comparing to the input dimension. The intuition of the linear speedup of the asynchronous proximal algorithms with variance reduction can be explained as follows. Asynchronous implementation updates the master parameter based on the delayed gradients.  If the data are sparse for asynchronous ProxSVRG or the coordinate block size is small comparing to the input dimension for ProxSVRCD, the influence of the delayed gradients can be bounded, and the asynchronous implementations are roughly equivalent to the sequential version.

In addition to the theoretical analysis, we have also conducted experiments on benchmark datasets to test the performances of the asynchronous stochastic proximal algorithms with variance reduction. According to the experimental results, we have the following observations: (1) Async-ProxSVRG have good speedup, especially for sparse data; (2) Async-ProxSVRCD also have good speedup, and is more efficient than Async-ProxSVRG when the input feature vectors are relatively dense or the coordinate block size is small. (3) Async-ProxSVRG and Async-ProxSVRCD can converge faster than other asynchronous algorithms reported in literature such as Async-ProxSGD \cite{lian2015asynchronous} and Async-ProxSCD \cite{liu2015asynchronous}. The results are consistent across different datasets, indicating that our observations are general and the two asynchronous proximal algorithms are highly efficient and scalable for practical use.

This paper is organized as follows: in Section 2, we briefly introduce the stochastic proximal algorithms with variance reduction including ProxSVRG and ProxSVRCD, and then related works; in Section 3, we describe the asynchronous parallelization of these algorithms; in Section 4, we prove the convergence rates for Async-ProxSVRG and Async-ProxSVRCD; in Section 5, we report the experimental results and make discussions; finally, in the last section, we conclude the paper and present future research directions.

\section{Background}
In this section, we will briefly introduce proximal algorithms with variance reduction, and then review the existing convergence analysis for asynchronous parallel algorithms.
\subsection{ProxSGD and ProxSCD}
At first, let us briefly introduce the standard stochastic proximal gradient algorithms,i.e., ProxSGD and ProxSCD. With ProxSGD, at iteration $k$, the solution to the R-ERM problem (i.e., Eqn (\ref{eq1})) is as follows:
{\small\begin{eqnarray}\label{eq2}
	&&x_{k+1}={prox_{\eta_kR}}\left\{x_{k}-\eta_k\nabla f_{\mathcal{B}_k}(x_{k}))\right\},
	\end{eqnarray}}
where $\eta_k$ is the step size, $\mathcal{B}_k$ is a mini-batch of randomly selected training data, $\nabla f_{\mathcal{B}_k}(x_k)=\frac{1}{|\mathcal{B}_k|}\sum_{i\in\mathcal{B}_k}\nabla f_i(x_k)$ and the proximal mapping is defined as $prox_R(y)=argmin_{x\in \mathbb{R}^d}\left\{\frac{1}{2}\|x-y\|_2^2+R(x)\right\}$.

ProxSCD exploits the block separability of the regularization term $R$ in the R-ERM problem, i.e.,$R(x_\cdot)=\sum_{j=1}^{m}R_j(x_{\cdot,C_j})$, where $x_{\cdot,C_j}$ is the $j$-th coordinate block of $x_{\cdot}$.
For example, for the $L1$-norm regularizer, $\{C_j;j=1,\cdots,m\}$ is a partition of $\{1,\cdots,d\}$ with $m=\frac{d}{block\ size}$, and $R_j(x_{\cdot,C_j})=\sum_{l\in C_j}|x_{\cdot,j}|$. ProxSCD randomly selects a coordinate block and update the coordinates in that block based on their gradients while keep the value of the other coordinates unchanged, i,e.,
{\small\begin{eqnarray}\label{eq5}
	&&x_{k+1,C_{j_k}}={prox_{\eta R_{j_k}}}\left\{x_{k,C_{j_k}}-\eta\nabla_{C_{j_k}}F(x_{k-1}))\right\},
	\end{eqnarray}}
where $C_{j_k}$ is the coordinate block sampled at iteration $k$, and $\nabla_{C_{j}}F(x)=[\nabla F(x)]_{C_j}$.
\subsection{Proximal Algorithms with Variance Reduction}

For ProxSGD, the step size $\eta_k$ has to be decreasing in order to mitigate the variance introduced by random sampling, which usually leads to slow convergence. To tackle this problem, one of the most popular variance reduction techniques was proposed by Johnson and Zhang \cite{johnson2013accelerating}. Xiao and Zhang applied this variance reduction technique to improve ProxSGD, and a new algorithm called ProxSVRG was proposed \cite{xiao2014proximal}.

The ProxSVRG algorithm divides the optimization process into multiple stages. At the beginning of stage $s$, ProxSVRG calculates the full gradient at the current solution $\tilde{x}_{s-1}$, i.e., $ \nabla F(\tilde{x}_{s-1})$. Then, at iteration $k$ inside stage $s$, the solution is updated as follows:
{\small\begin{eqnarray}
	v_k&=&\nabla f_{\mathcal{B}_k}(x_{k})-\nabla f_{\mathcal{B}_k}(\tilde{x}_{s-1})+\nabla F(\tilde{x}_{s-1}), \label{eq3}\\
	x_{k+1}&=&{prox_{\eta_kR}}\left\{x_{k}-\eta_kv_k\right\}, \label{eq4}
	\end{eqnarray}}
where $-\nabla f_{\mathcal{B}_k}(\tilde{x}_{s-1})+\nabla F(\tilde{x}_{s-1})$ is the variance reduction regularization term.

For ProxSCD, since the variance introduced by the block selection asymptotically goes to zero, it attains linear convergence rate. However, it still requires that all component functions are accessible within every iteration. Zhao $et.al.$ used variance reduction technique to improve ProxSCD with random training data sampling and a new algorithm called ProxSVRCD was proposed \cite{zhao2014accelerated}. \footnote{In \cite{zhao2014accelerated}, this algorithm was named MRBCD. In this paper, we call it ProxSVRCD to ease our reference.}

ProxSVRCD is similar to ProxSVRG, the update formula for iteration $k$ inside stage $s$ takes the following form:
{\small\begin{align}
	&v_{k}=\nabla f_{\mathcal{B}_k}(x_{k})-\nabla f_{\mathcal{B}_k}(\tilde{x}_{s-1})+\nabla F(\tilde{x}_{s-1}),\label{eq9}\\
	&x_{k+1,C_{j_k}}={prox_{\eta_kR_{j_k}}}\left\{x_{k,C_{j_k}}-\eta_k v_{k,C_{j_k}}\right\}, \label{eq10}\\
	&x_{k+1,\setminus C_{j_k}}\leftarrow x_{k,\setminus C_{j_k}}.\label{eq11}
	\end{align}}
where {\small$-\nabla f_{\mathcal{B}_k}(\tilde{x}_{s-1})+\nabla F(\tilde{x}_{s-1})$} is the variance reduction regularization term.

\subsection{Existing Convergence Analysis of Asynchronous Parallel Algorithms}
The asynchronous parallel methods have been successfully applied to accelerate many optimization algorithms including stochastic gradient descent (SGD)\cite{agarwal2011distributed,feyzmahdavian2015asynchronous,recht2011hogwild,mania2015perturbed}, stochastic coordinate descent (SCD) \cite{liu2013asynchronous,liu2015asynchronous}, stochastic dual coordinate ascent (SDCA) \cite{tran2015scaling} and randomized Kaczmarz algorithm \cite{liu2014asynchronous}. However, to the best of our knowledge, the asynchronous parallel versions of ProxSVRG and ProxSVRCD are not well studied, as well as their theoretical properties.

We briefly review the works which are closely related to ours as follows.
Reddi $et.al.$ studied asynchronous SVRG and proved that, asynchronous SVRG can achieve near linear speedup under some sparse condition \cite{reddi2015variance}.
Liu and Wright analyzed the asynchronous ProxSCD. They proved that the asynchronous ProxSCD can achieve near linear speedup if the delay is bounded by $\mathcal{O}(d^{\frac{1}{4}})$, where $d$ is the input dimension \cite{liu2015asynchronous}.

However, to the best of our knowledge, there is no study on the asynchronous parallel versions of proximal algorithms with variance reduction, as well as their theoretical properties.

\section{Asynchronous Proximal Algorithms with Variance Reduction }
In this section, we describe our Async-ProxSVRG and Async-ProxSVRCD algorithms under the following asynchronous parallel architecture. Suppose there are $P$ local workers and one master. For local workers, each of them has full access to the training data and stores a non-overlapping partition $N_p$ $(p=1,...,P)$ of the training data. Each local worker independently communicates with the master to \emph{pull} the global parameters from the master, and it computes the stochastic gradients locally and then \emph{push} the gradients to the master. For the master, it maintains the global model. It updates the model parameters with the gradient pushed by local workers and sends the model parameters to local workers when it receives the pull request. Master can control the access conflict based on different granularity. In Async-ProxSVRG, the local worker will access the entire model in every update. Therefore, we let master only response to one local worker's request at one time, which means the global model is atomic for all workers. In Async-ProxSVRCD, the local worker will only access a coordinate block in every update and different workers might work on different blocks without interfering others. In this case, master will response to multiple local workers simultaneously if only they are not accessing the same coordinate block, which means the global model is atomic at coordinate block level.

With variance reduction technique, the optimization process is divided into multiple stages (i.e., outer loop: $s=1,\cdots,S$). In each stage, there are two phases: full gradient computation and solution updates (i.e., inner loop: $k=1,\cdots, K$).

\emph{Full gradient computation}: the workers collectively compute the full gradient in parallel based on the entire training data. Specifically, each worker pulls the master parameter from the master, computes the gradients over one part of the training data, and pushes the sum of the gradients to the master. Then the master aggregates the gradients from the workers to obtain the full gradient, and broadcasts it to the workers.

\emph{Solution updates}: the workers compute the VR-regularized stochastic gradient in an asynchronous way and the master makes updates according to the proximal algorithms. To be specific, at iteration $k$, one local worker (who just finished its local computation) pulls the master parameters from the master, computes the VR-regularized stochastic gradient according to Eqn (\ref{eq3}) for ProxSVRG or Eqn(\ref{eq9}) for ProxSVRCD, and then pushes it to the master without any synchronization with the other workers. After the master receives the VR-regularized gradient from this worker, it updates the master parameter according to Eqn (\ref{eq4}) for ProxSVRG or Eqn (\ref{eq10})(\ref{eq11}) for Prox SVRCD. Then the global clock becomes $k+1$, and the next iteration begins. Corresponding details can be found in Algorithm \ref{Alg1}.

Please note that, the gradient pushed by a local worker to the master could be delayed. The reason is, when the worker is working on its own local computation, other workers might finish their computations and push their gradients to the master, and the master updates the master parameter accordingly.

As aforementioned, for Async-ProxSVRG, the whole model is atomic to each worker’s access. When the worker $0$ is working on its own local computation, worker $1$ and worker $2$ might  finish their computations, pushed their gradients to the master, and the master updates the master parameter accordingly. Thus, when worker $0$ finish its computation and push it to the master, the global clock has already plus $2$. Thus, the local gradients have delay=$2$ for the current master parameter. We use a random variable $\tau_k$ to denote the \emph{delay} of local gradients received by the master at global clock $k$. The delay equals to the number of updates that other workers have committed to the master between one particular worker pulls the parameter from the master and pushes gradients to the master.
For asynchronous ProxSVRCD, multiple workers may access the master parameter simultaneously, updating different coordinate blocks. Then different coordinate blocks in the model could be inconsistent regarding to the global update clock. To be precise, at global clock $k$, the master makes update based on the gradients computed by a local worker, who read the first coordinate block of the master parameter at global clock $k-\tau_k$. We denote the finally pulled parameter as $\hat{x}_k$, which can be represented as below:
{\small\begin{equation}\label{incon}
	\hat{x}_k=x_{k-\tau_k} + \sum_{h\in J(k)}(x_{h+1}-x_h),
	\end{equation}}
where {\small$J(k)\subset\{k-\tau_k,…,k-1\}$}. The $k$-th update can be described as {\small$x_{k+1,C_{j_k}}={prox_{\eta_kR_{j_k}}}\left\{x_{k,C_{j_k}}-\eta_k u_{k,C_{j_k}}\right\},$} where {\small$u_{k}=\nabla f_{\mathcal{B}_k}(\hat{x}_k)-\nabla f_{\mathcal{B}_k}(\tilde{x})+\nabla F(\tilde{x})$}. The delay $\tau_k$ equals to the difference between the clock at which a local worker pulls the first coordinate block from the master and the clock at which the local worker pushes the gradients to the master.

We conduct theoretical analysis for Async-ProxSVRG and Async-ProxSVRCD based on the above setting in the next section. Like other asynchronous parallel algorithms, the delay also plays an important role in the convergence rate of asynchronous proximal algorithms with variance reduction.
\begin{algorithm}[!ht]
	\caption{Async-ProxSVRG and Async-ProxSVRCD}
	\label{Alg1}
	\begin{algorithmic}
		\REQUIRE initial vector $\tilde{x}_0$, step size $\eta$, number of inner loops $K$, size of mini-batch $B$, number of coordinate blocks $m$.
		\ENSURE $\tilde{x}_S$
		\FOR{$s=1,2,...,S$}
		\STATE {\small$\tilde{x}=\tilde{x}_{s-1}$, $x_0=\tilde{x}$}
		\STATE \emph{For local worker $p$:} calculate {\small$\nabla F_p(\tilde{x})=\sum_{i\in N_p}\nabla f_i(\tilde{x})$} and send it to the master.
		\STATE \emph{For master:} calculate {\small$\nabla F(\tilde{x})=\frac{1}{n}\sum_{p=1}^{P}\nabla F_p(\tilde{x})$} and send it to each local worker.
		\FOR{$k=1,...,K$}
		\STATE \textbf{1. Async-ProxSVRG: consistent read}
		\STATE \emph{For local worker p:} randomly select a mini-batch $\mathcal{B}_k$ with $|\mathcal{B}_k|=B$.
		\STATE \textbf{Pull} current state $x_{k-\tau_k}$ from the master.
		\STATE \textbf{Compute} {\small$u_{k}=\nabla f_{\mathcal{B}_k}(x_{k-\tau_k})-\nabla f_{\mathcal{B}_k}(\tilde{x})+\nabla F(\tilde{x})$.}
		\STATE \textbf{Push} $u_{k}$ to the master.
		\STATE \emph{For master:}
		\STATE \textbf{Update} {\small$x_{k+1}=prox_{\eta R}(x_{k}-\eta u_{k})$.}
		\STATE \textbf{2. Async-ProxSVRCD: inconsistent read}
		\STATE \emph{For local worker p:} randomly select $\mathcal{B}_{k}$ with $|\mathcal{B}_k|=B$, and randomly select $j_k\in[m]$.
		\STATE \textbf{Pull} current state $\hat{x}_{k}$ from the master.
		\STATE \textbf{Compute} {\small$u_{k}=\nabla f_{\mathcal{B}_k}(\hat{x}_k)-\nabla f_{\mathcal{B}_k}(\tilde{x})+\nabla F(\tilde{x})$}.
		\STATE \textbf{Push} $u_{k}$ to the master.
		\STATE \emph{For master:}
		\STATE \textbf{Update} {\small$x_{k+1,C_{j_k}}={prox_{\eta R_{j_k}}}\left\{x_{k,C_{j_k}}-\eta u_{k,C_{j_k}}\right\}$};\\
		\quad\quad\quad\ {\small$x_{k+1,\setminus C_{j_k}}\leftarrow x_{k,\setminus C_{j_k}}$}
		\ENDFOR
		\STATE {\small$\tilde{x}_{s}=\frac{1}{K}\sum_{k=1}^{K}x_k$}
		\ENDFOR
	\end{algorithmic}
\end{algorithm}		
\section{Convergence Analysis}
In this section, we prove the convergence rates of the asynchronous parallel proximal algorithms with variance reduction introduced in the previous section.

\subsection{Async-ProxSVRG}
At first, we introduce the following assumptions, which are very common in the theoretical analysis for asynchronous parallel algorithms \cite{recht2011hogwild,reddi2015variance}.

\textbf{Assumption 1}: (Convexity) $F(x)$ and $R(x)$ are convex and $R(x)$ is block sparable. The objective function $P(x)$ is $\mu$-strongly convex, i.e., $\forall x,y\in \mathbb{R}^d$, we have,	
{\small\begin{equation*}
	P(y)\geq P(x)+\xi^T(y-x)+\frac{\mu}{2}\|y-x\|^2, \forall\xi\in \partial P(x).\footnote{In this paper, if there is no specification, $\|\cdot\|$ is the $L_2$-norm.}
	\end{equation*}}
\textbf{Assumption 2}: (Smoothness)
The components $\{f_i(x);i\in[n]\}$ of $F(x)$ are differentiable and have Lipschitz continuous partial gradients and thus Lipschitz continuous gradients, i.e., $\exists T,L>0$, such that  $\forall x,y\in \mathbb{R}^d$ with $x_j\neq y_j$, we have
{\small\begin{align*}
	\|\nabla_jf_i(x)-\nabla_jf_i(y)\|&\leq T\|x_j-y_j\|,\forall i\in[n], j\in[d].\\
	\|\nabla f_i(x)-\nabla f_i(y)\|&\leq L\|x-y\|, \forall i.
	\end{align*}}		
\textbf{Assumption 3}: (Bounded and Independent Delay) The random delay variables $\tau_1, \tau_2,...$ in consistent read setting are independent of each other and independent of $\mathcal{B}_k$, and their expectations are upper bounded by $\tau$, i.e., $\mathbb{E}\tau_k\leq\tau$ for all $k$.\\	
\textbf{Assumption 4}: (Data Sparsity) The maximal frequency of a feature appearing in the dataset is upper bounded by $\Delta$.

Based on these assumptions, we prove that Async-ProxSVRG has linear convergence rate. 	
\begin{theorem}\label{thm3.1}
	Suppose Assumptions 1-4 hold. If the step size  $\eta<\min\left\{\frac{2}{5LB\Delta\tau^2},\frac{B}{16L}\right\}$, and the inner loop size $K$ is sufficiently large so that
	{\small\begin{equation*}
		\rho=\frac{B}{\eta\mu K(B-8\eta L)}+\frac{8\eta L}{(B-8\eta L)}<1,
		\end{equation*}}
	then Async-ProxSVRG has linear convergence rate in expectation:
	{\small\begin{equation*}
		\mathbb{E}P(\tilde{x}_s)-P(x^*)\leq \rho^s[P(\tilde{x}_0)-P(x^*)],
		\end{equation*}}
	where $x^*=argmin_xP(x)$.
\end{theorem}
Due to space limitation, we only provide the proof sketch and put the proof details into supplementary materials.\\
\textbf{Proof Sketch of Theorem \ref{thm3.1}:}

Firstly we introduce some notations.
Let $x_{k+1}-x_{k}=-\eta g_k$, $v_{k}=\nabla f_{\mathcal{B}_k}(x_{k})-\nabla f_{\mathcal{B}_k}(\tilde{x})+\nabla F(\tilde{x})$, and $u_{k}=\nabla f_{\mathcal{B}_k}(x_{k-\tau_k})-\nabla f_{\mathcal{B}_k}(\tilde{x})+\nabla F(\tilde{x})$.\\

\textit{Step 1:}
The key for the proof is that by the spasity condition, we have
{\small$
	F(x)\geq F(y)-\nabla F(x)(y-x)-\frac{LB\Delta}{2}\|x-y\|^2.$
}

\textit{Step 2:}
By using the convexity of $F(x)$ and $R(x)$, we have:
{\small\begin{eqnarray*}
		&& P(x^*)\geq P(x_{k+1})+(u_k-\nabla F(x_{k-\tau_k}))^T(x_{k+1}-x^*)\\
		&&\quad\quad+\eta\|g_k\|^2-\frac{L\eta^2B\Delta\tau_k}{2}\sum_{h=k-\tau_k}^{k}\|g_h\|^2+g_k^T(x^*-x_{k+1}).
	\end{eqnarray*}}
	\textit{Step 3:}
	We use Lemma 3 in \cite{xiao2014proximal} to bound the term $\mathbb{E}_{\mathcal{B}_k}(u_k-\nabla F(x_{k-\tau_k}))^T(x_{k+1}-x^*)$. Then by summing $k$ from $0$ to $K-1$, we can get:
	{\small
		$-\sum_{k=0}^{K-1}\mathbb{E}g_k^T(x_{k}-x^*)+\left(\eta-L\eta^2B\Delta\tau^2(1+2L\eta)\right)\sum_{k=0}^{K-1}\mathbb{E}\|g_k\|^2
		\leq(\frac{8L\eta}{B}-1)\sum_{k=0}^{K-1}(P(x_{k+1})-P(x^*))
		+\frac{8L\eta}{B}(K+1)(P(\tilde{x})-P(x^*)).
		$}		
	
	\textit{Step 4:}
	Under the condition $\eta<\min\left\{\frac{2}{5LB\Delta\tau^2},\frac{B}{16L}\right\}$, we have
	$\eta-L\eta^2B\Delta\tau^2(1+2L\eta)\geq \frac{\eta}{2}$. Then following the proof of ProxSVRG, we can get the results.
	
	\textbf{Remark}: Theorem \ref{thm3.1} actually shows that, Async-ProxSVRG can achieve linear speedup when $\Delta$ is small and $\tau\leq\sqrt{8/B^2\Delta}$. 
	For sequential ProxSVRG, with step size $\eta=0.1B/L$, the inner loop size $K$ should be in the same order of $\mathcal{O}(L/B\mu)$ to make $\rho<1$. The computation complexity (number of gradients need to calculate) for the inner loop is in the same order of $\mathcal{O}(L/\mu)$.
	For the Async-ProxSVRG, with $\eta=\min\{\frac{2}{5LB\Delta\tau^2},\frac{0.05B}{L}\}$, the inner loop size $K$ should be in the same order of $\mathcal{O}(L/B\mu+B\Delta\tau^2L/\mu)$ to make $\rho<1$. For the case $\tau<\sqrt{8/B^2\Delta}$ (i.e., $\frac{0.05B}{L}<\frac{2}{5LB\Delta\tau^2}$), by setting $\eta=\frac{0.05B}{L}$, the order of inner loop size $K$ is $\mathcal{O}(L/B\mu)$ and the corresponding computation complexity is {\small$\mathcal{O}(L/\mu)$}, which is the same as the sequential ProxSVRG. Therefore, Async-ProxSVRG can achieve nearly the same performance as the sequential version, but $\tau$ times faster since we are running the algorithm asynchronously, and thus we achieve "linear speedup". For the case $\tau\geq\sqrt{8/B^2\Delta}$, the inner loop size $K$ should be in the same order of $\mathcal{O}(B\Delta\tau^2L/\mu)$. Compared with the sequential ProxSVRG with $K=\mathcal{O}(L/B\mu)$, Async-ProxSVRG can not obtain linear speedup but still have a theoretical speedup of $1/B^2\Delta\tau$ if $B^2\Delta\tau<1$.
	
	According to Theorem \ref{thm3.1} and the above discussions, we provide the following corollary for a simple setup of the parameters in Async-ProxSVRG which can achieve near linear speedup.
	\begin{corollary}\label{coro3.2}
		Suppose Assumptions 1-4 hold. If we set $B=\left(\frac{1}{\Delta}\right)^{\frac{1}{4}}$, $\tau\leq\sqrt{8/\Delta^{\frac{1}{2}}}$, $\eta=\frac{0.05\Delta^{\frac{1}{4}}}{L}$ and $K=\frac{200L\Delta^{\frac{1}{4}}}{\mu}$, then Async-ProxSVRG has the following linear convergence rate:
		{\small\begin{equation*}
			\mathbb{E}P(\tilde{x}_s)-P(x^*)\leq \left(\frac{5}{6}\right)^s[P(\tilde{x}_0)-P(x^*)],
			\end{equation*}}
		where $x^*=argmin_xP(x)$.
	\end{corollary}
	
	\subsection{Async-ProxSVRCD}
	In this section, we present Theorem \ref{thm3.5}, which states the convergence rate of Async-ProxSVRCD, as well as the conditions for them to achieve near linear speedup.
	
	\textbf{Assumption 3$'$}:(Bounded and Independent Delay) The random delay variables $\tau_1, \tau_2,...$ in inconsistent read setting in Eqn \ref{incon} are independent of each other and independent of $\mathcal{B}_k$, and their expectations are upper bounded by $\tau$.
	\begin{theorem}\label{thm3.5}
		Suppose Assumptions 1, 2, and 3$'$ hold. In addition, we assume that the mini-batch size $B\geq L/T$, the step size $\eta$ and the coordinate block number $m$ satisfies  {\small$\eta<\min\left\{\frac{1}{T}\frac{m^{\frac{3}{2}}-T\tau}{m^{\frac{3}{2}}+3m\tau+\tau^2},\frac{1}{8T},\frac{\mu\sqrt{m}}{2T\tau}\right\}$}, and the inner loop size $K$ is sufficiently large so that
		{\small\begin{equation*}
			\rho=\left(\frac{m}{\eta\mu K(1-\frac{T\eta\tau}{\mu\sqrt{m}}-4\eta T)}+\frac{4\eta T(K+1)}{(1-\frac{T\eta\tau}{\mu\sqrt{m}}-4\eta T)K}\right)<1,
			\end{equation*}}
		then Async-ProxSVRCD has linear convergence in expectation:
		\begin{equation*}
		\mathbb{E}P(\tilde{x}_s)-P(x^*)\leq \rho^s[P(\tilde{x}_0)-P(x^*)],
		\end{equation*}
		where $x^*=argmin_xP(x)$.
	\end{theorem}
	\textbf{Proof Sketch of Theorem \ref{thm3.5}:}
	
	\textit{Step 1:} By the convexity of $F(x)$ and $R(x)$, we have
	{\small\begin{eqnarray}
		&& P(x^*)\geq m\mathbb{E}_{j_k}P(x_{k+1})-(m-1)P(x_{k})+(\eta-\frac{T\eta^2}{2})m\mathbb{E}_{j_k}\|g_k\|^2 \nonumber\\
		&&+(v_k-\nabla F(x_{k}))^T(\bar{x}_{k+1}-x^*)+m\mathbb{E}_{j_k}(g_k)^T(x^*-x_{k}) \nonumber\\
		&&+(\nabla f_{\mathcal{B}_k}(\hat{x}_k)-\nabla f_{\mathcal{B}_k}(x_{k}))^T(\bar{x}_{k+1}-x^*).\nonumber \\\nonumber
		\end{eqnarray}}
	\textit{Step 2:} We decompose the term {\small$-(\nabla f_{\mathcal{B}_k}(\hat{x}_k)-\nabla f_{\mathcal{B}_k}(x_{k}))^T(\bar{x}_{k+1}-x^*)$} by using Assumption 2 as below:
	{\small\begin{align*}
		&\quad-\frac{1}{T}(\nabla f_{\mathcal{B}_k}(\hat{x}_k)-\nabla f_{\mathcal{B}_k}(x_{k}))^T(\bar{x}_{k+1}-x^*)\leq  \nonumber\\
		& \sum_{a=k-\tau}^{k-1}\|x_{a+1}-x_{a}\|\|\bar{x}_{k+1}-x_{k}\|+\sum_{a=k-\tau}^{k-1}\|x_{a+1}-x_{a}\|\|x_{a}-x^*\|\\
		&+\sum_{a=k-\tau}^{k-1}\sum_{b=a}^{k-1}\|x_{a+1}-x_{a}\|\|x_{b+1}-x_{b}\|.
		\end{align*}}
	By taking expectation w.r.t $j_1,\cdots,j_k$ gradually, we can bound the three terms on the right side. This is a key step for the proof and please see the details in the supplementary materials. Thus we can get
	{\small\begin{align*}
		&\quad\sum_{k=1}^{K}-\mathbb{E}(g_k)^T(x_{k}-x^*)+A(\eta)\mathbb{E}\|g_k\|^2\\
		&\leq \sum_{k=1}^{K}-\left(\mathbb{E}P(x_{k+1})-P(x^*)\right)-\frac{1}{m}(v_k-\nabla F(x_{k}))^T(\bar{x}_{k+1}-x^*)\\
		&\quad+\sum_{k=1}^{K}(\frac{(m-1)}{m}+\frac{T\eta\tau}{\mu m^{\frac{3}{2}}})\mathbb{E}\left(P(x_{k})-P(x^*)\right),
		\end{align*}}
	where $A(\eta)=\left(\eta(1-\frac{T\tau}{2m^{\frac{3}{2}}})-T\eta^2(\frac{1}{2}+\frac{\tau}{2\sqrt{m}}+\frac{\tau}{m}+\frac{\tau^2}{2m^{\frac{3}{2}}})\right)$.
	\textit{Step 3:}		
	With the assumption {\small$\eta<\frac{1}{T}\frac{m^{\frac{3}{2}}-T\tau}{m^{\frac{3}{2}}+3m\tau+\tau^2}$}, we have {\small$A(\eta)>\frac{\eta}{2}$}. Then by following the proof of ProxSVRCD, we can get the results.
	
	\textbf{Remark:} Theorem \ref{thm3.5} actually shows that when $m$ is large (or equivalent the block size is small) and {\small$\tau\leq\min\left\{\sqrt{m},4\mu\sqrt{m},m^{\frac{3}{2}}/2T\right\}$}, Async-ProxSVRCD can achieve linear speedup. 
	For the sequential ProxSVRCD, Corollary 4.3 in \cite{zhao2014accelerated} set $\eta=1/16T$, $B=L/T$ and the inner loop size $K$ in the same order of $\mathcal{O}(mT/\mu)$ to make $\rho<1$. For Async-ProxSVRCD, if $m$ is sufficiently large so that the delay satisfies {\small$\tau\leq\min\left\{\sqrt{m},4\mu\sqrt{m},m^{\frac{3}{2}}/2T\right\}$}, we can set $\eta=1/24T$ which guarantees the condition {\small$\eta<\min\left\{\frac{1}{T}\frac{m^{\frac{3}{2}}-T\tau}{m^{\frac{3}{2}}+3m\tau+\tau^2},\frac{1}{8T},\frac{\mu\sqrt{m}}{2T\tau}\right\}$}. Thus, the inner loop size $K$ should be $\mathcal{O}(mT/\mu)$ to make $\rho<1$, which is the same as sequential ProxSVRCD. Therefore, Async-ProxSVRCD can achieve near linear speedup. If we consider the  indicative case \cite{shamir2014communication} in which $L/\mu=\sqrt{n}$, $L=\mathcal{O}(1)$ and $\mu=\mathcal{O}(\sqrt{1/n})$. The condition for the linear speedup can be simplified to $\tau\leq4\sqrt{m/n}$. Even if $4\sqrt{m/n}<\tau\leq\sqrt{m}$, Async-ProxSVRCD still have a speedup of {\small$\mathcal{O}(\sqrt{m/n})$} by setting {\small$\eta\leq\frac{\mu\sqrt{m}}{2T\tau}=\frac{\sqrt{m}}{2\tau\sqrt{n}}$}, since{\small$\frac{m^{\frac{3}{2}}-T\tau}{m^{\frac{3}{2}}+3m\tau+\tau^2}>\frac{\sqrt{m}}{2\tau\sqrt{n}}$}.
	
	According to Theorem \ref{thm3.5} and the above discussions, we provide the following corollary for a simple setup of the parameters in Async-ProxSVRCD which can achieve near linear speedup.
	\begin{corollary}\label{coro3.4}
		Suppose Assumptions 1,2, and 3$'$ hold and the delay bound satisfies $\tau\leq\min\left\{\sqrt{m},4\mu\sqrt{m},m^{\frac{3}{2}}/2T\right\}$. Let $\eta=1/24T$, $B=L/T$ and $K=\frac{216mT}{\mu}$, then Async-ProxSVRCD has the following linear convergence rate:
		{\small\begin{equation*}
			\mathbb{E}P(\tilde{x}_s)-P(x^*)\leq \left(\frac{5}{6}\right)^s[P(\tilde{x}_0)-P(x^*)],
			\end{equation*}}
		where $x^*=argmin_xP(x)$.
	\end{corollary}

	By comparing the conditions of the linear speedup for asynchronous Proximal algorithms, we have the following findings: (1). Async-ProxSVRG relies on the data sparsity to alleviate the negative impact of communication delay $\tau$; (2) Async-ProxSVRCD does not rely on the sparsity condition, however, it requires the block size is small or the input dimension is large, since in this way, the block-wise updates will become frequent and can also alleviate the delay of the whole parameter vector.
	
	To sum up, in this section, based on a few widely used assumptions, we have proven the convergence properties of the asynchronous parallel implementations of ProxSVRG, and ProxSVRCD, and discussed the conditions for them to achieve near linear speedups as compared to their sequential (single-machine) counterparts. In the next section, we will report the results of our experiments to verify these theoretical findings.

	\section{Experiments}
	In this section, we report our experimental results on the efficiency of the asynchronous proximal algorithms with variance reduction.
	In particular, we conducted binary classifications on three benchmark datasets: \emph{rcv1, real-sim, news20} \cite{reddi2015variance}, \emph{new20} is the densest one with a much higher dimension and \emph{rcv1} is the sparsest one. The detailed information about the three data sets is given in Table \ref{tab:1}. We use the logistic loss function with both $L_1$ and $L_2$ regularizations with weight $\lambda_1$ and $\lambda_2$ respectively.
	\begin{table}[h]
		\centering
		\caption{Experimental Datasets}\label{tab:1}
		\begin{tabular}{lllll}
			\hline\hline
			Dataset&\emph{rcv1}&\emph{real-sim}&\emph{news20}\\
			\hline
			Data size $n$ &20242&72309&19996\\
			Feature size $d$ &47236&20958&1355191\\
			$\lambda_1,\lambda_2$ &$10^{-5},10^{-4}$&$10^{-4},10^{-4}$&$10^{-6},10^{-4}$\\
			\hline\hline
		\end{tabular}
	\end{table}
	
	\begin{figure}[h]
		\centering
		\subfigure[Async-ProxSVRG]{
			\label{figa}
			\includegraphics[width=1.6in]{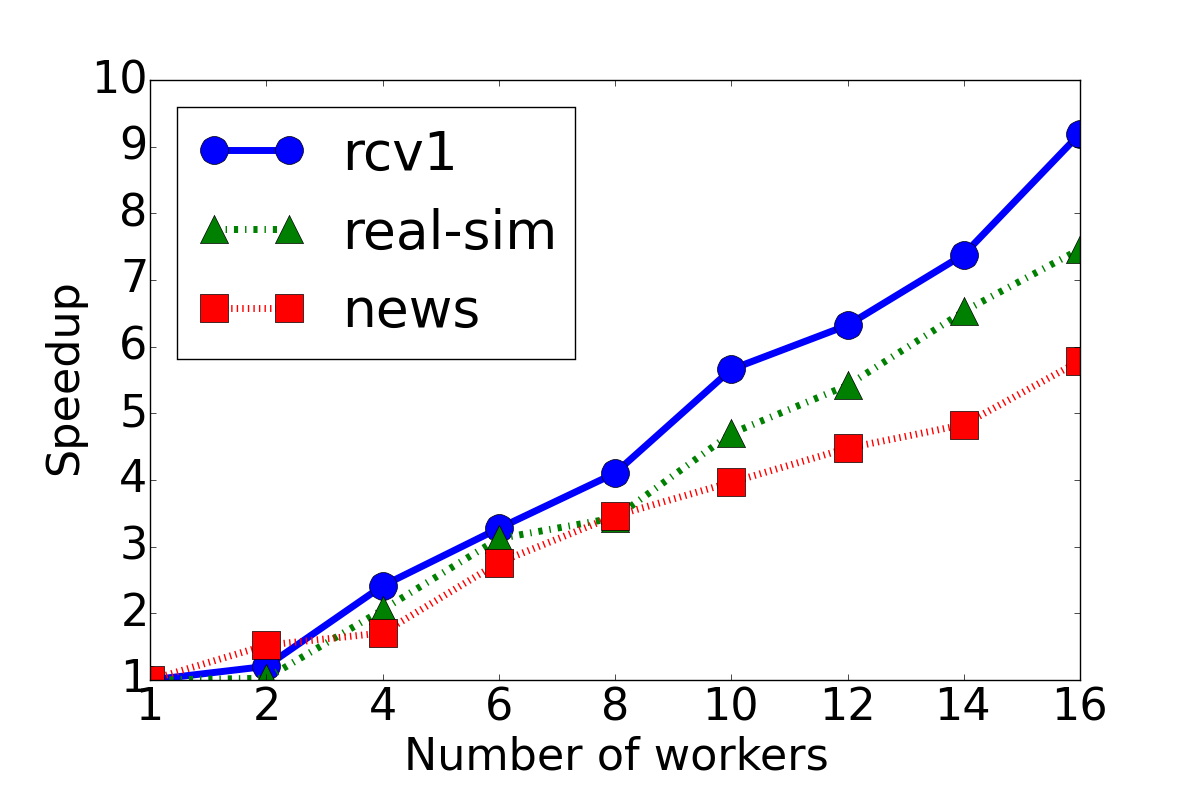}}
		\subfigure[Async-ProxSVRCD]{
			\label{figb}
			\includegraphics[width=1.6in]{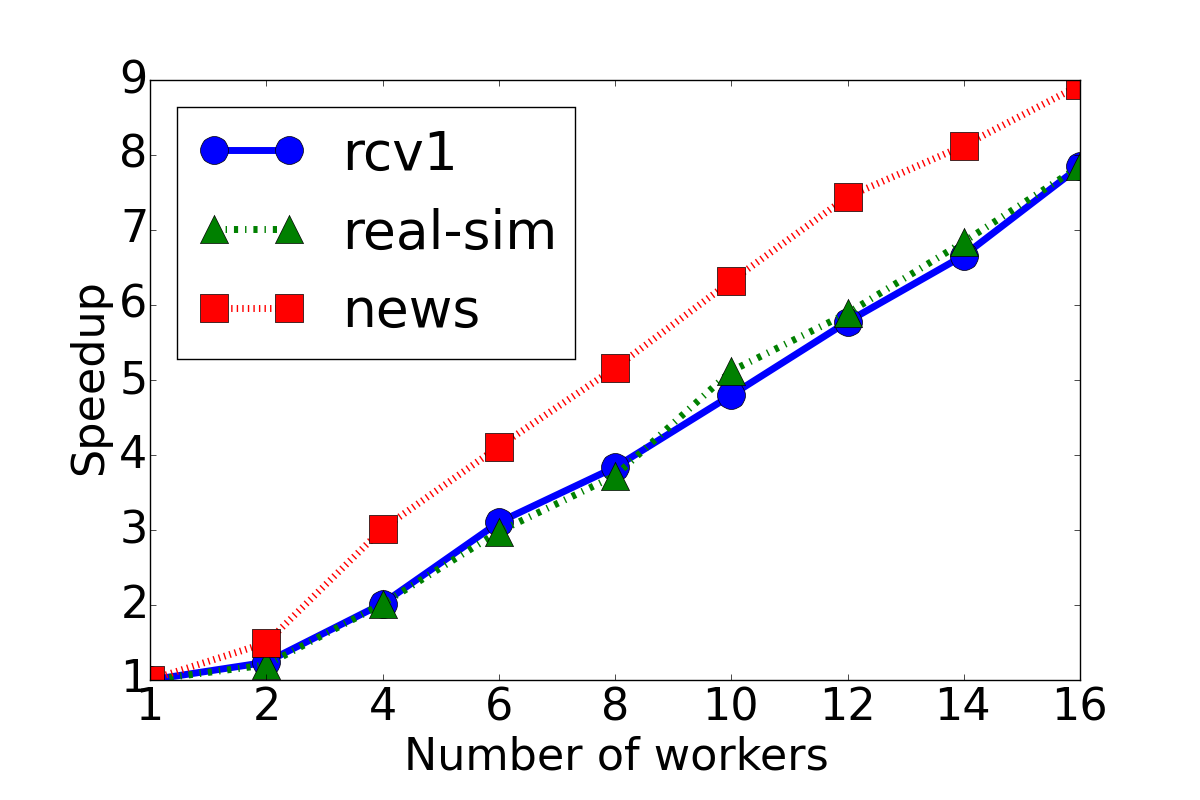}}
		\caption{Results for the speedups of asynchronous algorithms}
	\end{figure}
	
	Following the practices in \cite{xiao2014proximal}, we normalized the input vector of each data set before feeding it into the classifier, which leads to an upper bound of $0.25$ for the Lipschitz constant $L$. The stopping criterion for all the algorithms under investigation is the optimization error smaller than $10^{-10}$ (i.e., $P(\tilde{x}_S)-P(x^*)<10^{-10}$). 	
	For Async-ProxSVRG, we set step size $\eta=0.04$, the mini-batch size $B=200$, and the inner loop size $K=2n$, where $n$ is the data size. For Async-ProxSVRCD, we set step size $\eta=0.04$, the number of block partitions $m=\frac{d}{100}$, the mini-batch size $B=200$, and a larger inner loop size $K=2nm$. We implement Async-ProxSVRG and Async-ProxSVRCD in the consistent read setting and the inconsistent read setting, respectively.
	
	The speedups of Async-ProxSVRG and Async-ProxSVRCD are shown in Figures \ref{figa} and \ref{figb}. From the figures, we have the following observations. (1) On all the three datasets, Async-ProxSVRG has near linear speedup compared to its sequential counterpart. The speedup on \emph{rcv1} is the largest, while that on \emph{news20} is the smallest. This observation is consistent with our theoretical findings that Async-ProxSVRG has better performance on sparser data. (2) Async-ProxSVRCD also achieves nice speedup. The speedup is more significant for \emph{news20} than that for the other two data sets. This is consistent with our theoretical discussions - the sufficient condition for the linear speedup of Async-ProxSVRCD is easier to be satisfied for high-dimensional datasets.
	As literature also reported other asynchronous algorithms,
	such as Async-ProxSGD and Async-ProxSCD, we also compare with them to test the performance
	of our algorithms. Our algorithms actually converge faster than those without
	variance reduction, which means asynchronization can work together with VR
	techniques smoothly and enhances the model's convergence speed. For saving space,
	we put the detailed results in the supplementary materials.
	
	In summary, our experimental results well validate our theoretical findings, and indicate that the asynchronous proximal algorithms with variance reduction are very efficient and could have good applications in practice.
	
	\section{Conclusion}
	In this paper, we have studied the asynchronous parallelization of two widely used proximal gradient algorithms with variance reduction, i.e., ProxSVRG and ProxSVRCD. We have proved their convergence rates, discussed their speedups, and verified our theoretical findings through experiments.  Overall speaking, these asynchronous proximal algorithms can achieve linear speedup under certain conditions, and can be highly efficient when being used to solve large scale R-ERM problems.  As for future work, we plan to make the following explorations. First, we will extend the study in this paper to the non-convex case, both theoretically and experimentally. Second, we will study the asynchronous parallelization of more proximal algorithms.

	\bibliographystyle{aaai}
	\bibliography{ASVRG}
			
			\newpage
			\onecolumn

		\section{Appendices}
		\subsection{Proof of Theorem 4.1}	
	
		Firstly we introduce some notations.
		Let $x_{k+1}-x_{k}=-\eta g_k$, $v_{k}=\nabla f_{\mathcal{B}_k}(x_{k})-\nabla f_{\mathcal{B}_k}(\tilde{x})+\nabla F(\tilde{x})$, and $u_{k}=\nabla f_{\mathcal{B}_k}(x_{k-\tau_k})-\nabla f_{\mathcal{B}_k}(\tilde{x})+\nabla F(\tilde{x})$.\\
		Since the update formula is
		{\small\begin{equation*}
			x_{k+1}=argmin_{x\in R^d}\left\{\frac{1}{2}\|x-x_{k}-\eta u_k\|^2+\eta R(x)\right\},	
			\end{equation*}}
		the associated optimality condition states that there is a $\xi_{k+1}\in \partial R(x_{k+1})$ such that
		{\small$x_{k+1}-(x_{k}-\eta u_k)+\eta\xi_{k+1}=0$}.
		By the smoothness assumption of $f_{i}(x)$ and the sparseness assumption 4, we have: $\forall x,y\in R^d$ which are independent with $\mathcal{B}_k$,
		{\small\begin{align}
			F(x)&=\mathbb{E}_{\mathcal{B}_k}f_{\mathcal{B}_k}(x)\nonumber\\&\geq \mathbb{E}_{\mathcal{B}_k}f_{\mathcal{B}_k}(y)-\mathbb{E}_{\mathcal{B}_k}\nabla f_{\mathcal{B}_k}(x)(y-x)-\frac{L}{2}\mathbb{E}_{\mathcal{B}_k}\|x-y\|_{\mathcal{B}_k}^2\nonumber\\
			&\geq F(y)-\nabla F(x)(y-x)-\frac{LB\Delta}{2}\|x-y\|^2,\label{ineq1}\\	\nonumber
			\end{align}}
		where Ineq.(\ref{ineq1}) is established by $\mathbb{E}_{\mathcal{B}_k}\|x\|_{\mathcal{B}_k}^2\leq B\Delta\|x\|^2$, which comes from $\mathbb{E}_{i_k}\|x\|_{i_k}^2\leq\Delta\|x\|^2$.
		
		By the convexity of $F(x)$ and $R(x)$, we have:
		{\small\begin{align}
			P(x^*)&=F(x^*)+R(x^*) \nonumber\\
			&\geq F(x_{k-\tau_k})+\nabla F(x_{k-\tau_k})^T(x^*-x_{k-\tau_k})+R(x_{k+1})+\xi_{k+1}^T(x^*-x_{k+1})\nonumber\\
			&\geq F(x_{k+1})-\nabla F(x_{k-\tau_k})^T(x_{k+1}-x_{k-\tau_k})-\frac{LB\Delta}{2}\|x_{k+1}-x_{k-\tau_k}\|^2+\nabla F(x_{k-\tau_k})^T(x^*-x_{k-\tau_k})\nonumber\\
			&\quad+R(x_{k+1})+\xi_{k+1}^T(x^*-x_{k+1}) \nonumber\\
			&=P(x_{k+1})+(u_{k}-\nabla F(x_{k-\tau_k}))^T(x_{k+1}-x^*)+g_k^T(x^*-x_{k+1})+\eta\|g_k\|^2-\frac{LB\Delta}{2}\|x_{k+1}-x_{k-\tau_k}\|^2 \nonumber\\
			&=P(x_{k+1})+(u_k-\nabla F(x_{k-\tau_k}))^T(x_{k+1}-x^*)+\eta\|g_k\|^2-\frac{L\eta^2B\Delta\tau_k}{2}\sum_{h=k-\tau_k}^{k}\|g_h\|^2+g_k^T(x^*-x_{k+1}), \label{ineq2}\\\nonumber
			\end{align}}
		where the second inequality is established by Ineq.(\ref{ineq1}).
		
		By rearranging Ineq.(\ref{ineq2}), we have: 				
		{\small\begin{align}
			&\quad-g_k^T(x_{k+1}-x^*)+(\eta-\frac{L\eta^2B\Delta\tau_k}{2})\|g_k\|^2 -\frac{L\eta^2B\Delta\tau_k}{2}\sum_{h=k-\tau_k}^{k}\|g_h\|^2 \nonumber\\
			&\leq P(x^*)-P(x_{k+1})-(u_k-\nabla F(x_{k-\tau_k}))^T(x_{k+1}-x^*) .\label{ineq3}\\\nonumber
			\end{align}}
		
		According to the proof of Lemma 3 in \cite{xiao2014proximal}, we can get:
		{\small\begin{align*}
			&\quad-\mathbb{E}_{\mathcal{B}_k}(u_k-\nabla F(x_{k-\tau_k}))^T(x_{k+1}-x^*)\\
			&=\eta \mathbb{E}_{\mathcal{B}_k}\|u_k-\nabla F(x_{k-\tau_k})\|^2\\
			&\leq 2\eta\mathbb{E}_{\mathcal{B}_k}\|(u_k-\nabla F(x_{k-\tau_k}))-(v_k-\nabla
			F(x_{k}))\|^2+2\eta\mathbb{E}_{\mathcal{B}_k}\|v_k-\nabla F(x_{k})\|^2\\
			&=2\eta\mathbb{E}_{\mathcal{B}_k}\|\nabla f_{\mathcal{B}_k}(x_{k-\tau_k})-\nabla f_{\mathcal{B}_k}(x_{k})\|^2-2\eta\|\nabla F(x_{k-\tau_k})-\nabla
			F(x_{k}))\|^2+2\eta\mathbb{E}_{\mathcal{B}_k}\|v_k-\nabla F(x_{k})\|^2\\
			&\leq 2\eta\mathbb{E}_{\mathcal{B}_k}\|\nabla f_{\mathcal{B}_k}(x_{k-\tau_k})-\nabla f_{\mathcal{B}_k}(x_{k})\|^2+2\eta\mathbb{E}_{\mathcal{B}_k}\|v_k-\nabla F(x_{k})\|^2\\
			&\leq 2\eta L^2\mathbb{E}_{\mathcal{B}_k}\|x_{k-\tau_k}-x_{k}\|_{\mathcal{B}_k}^2+2\eta\mathbb{E}_{\mathcal{B}_k}\|v_k-\nabla F(x_{k})\|^2\\
			&\leq 2\eta^3 L^2B\Delta\tau_k\sum_{h=k-\tau_k}^{k-1}\|g_h\|^2+2\eta\mathbb{E}_{\mathcal{B}_k}\|v_k-\nabla F(x_{k})\|^2\\
			&\overset{(1)}{\leq}2\eta^3 L^2B\Delta\tau_k\sum_{h=k-\tau_k}^{k-1}\|g_h\|^2+\frac{8\eta L}{B}[P(x_{k})-P(x^*)+P(\tilde{x})-P(x^*)].
			\end{align*}}
		The "$\overset{(1)}{\leq}$" is established based on Corollary 3 in \cite{xiao2014proximal}.
		
		Then by taking expectation on both sides of Ineq.(\ref{ineq3}) with respect to $\mathcal{B}_k$ and $\tau_k$, and by using Assumption 3, we obtain:
		{\small\begin{align}
			&\quad-\mathbb{E}_{\mathcal{B}_k}g_k^T(x_{k+1}-x^*)+(\eta-\frac{L\eta^2B\Delta\tau}{2})\mathbb{E}_{\mathcal{B}_k}\|g_k\|^2-(\frac{L\eta^2B\Delta\tau}{2}+2\eta^3 L^2B\Delta\tau)\sum_{h=k-\tau}^{k-1}\|g_h\|^2 \nonumber\\
			&\leq (P(x^*)-\mathbb{E}_{\mathcal{B}_k}P(x_{k+1}))+\frac{8L\eta}{B}(P(x_{k})-P(x^*)+P(\tilde{x})-P(x^*)) \label{ineq6}
			\end{align}}
		
		Summing both sides of Ineq.(\ref{ineq6}) from $k=0$ to $K-1$, and taking expectations with respect to $\mathcal{B}_{k-1},...,\mathcal{B}_{1}$ gradually, we can get:\\
		{\small\begin{align*}
			&-\sum_{k=0}^{K-1}\mathbb{E}g_k^T(x_{k+1}-x^*)+(\eta-\frac{L\eta^2B\Delta\tau}{2})\sum_{k=0}^{K-1}\mathbb{E}\|g_k\|^2-(\frac{L\eta^2B\Delta\tau}{2}+2\eta^3 L^2B\Delta\tau)\sum_{k=0}^{K-1}\sum_{h=k-\tau}^{k-1}\|g_h\|^2\\
			&\leq \sum_{k=0}^{K-1}(P(x^*)-\mathbb{E}P(x_{k+1}))+\frac{8L\eta}{B}\sum_{k=0}^{K-1}(P(x_{k})-P(x^*)+P(\tilde{x})-P(x^*)).\\
			\end{align*}}
		
		By reranging the above inequality, we can get:
		{\small\begin{align*}
			&\quad-\sum_{k=1}^{K-1}\mathbb{E}g_k^T(x_{k}-x^*)+\left(\eta-L\eta^2B\Delta\tau^2(1+2L\eta)\right)\sum_{k=0}^{K-1}\mathbb{E}\|g_k\|^2\\
			&\leq (\frac{8L\eta}{B}-1)\sum_{k=0}^{K-1}(P(x_{k+1})-P(x^*))+\frac{8L\eta}{B}(K+1)(P(\tilde{x})-P(x^*)).\\
			\end{align*}}
		
		Under the condition $\eta<\min\left\{\frac{2}{5LB\Delta\tau^2},\frac{B}{16L}\right\}$, we have
		$\eta-L\eta^2B\Delta\tau^2(1+2L\eta)\geq \frac{\eta}{2}$.
		Then we can get
		{\small\begin{align}
			\mathbb{E}\|x_K-x^*\|^2&=\mathbb{E}\|x_{K-1}-x^*\|^2-2\eta\mathbb{E}g_{K-1}^T(x_{K}-x^*)+\eta^2\mathbb{E}\|g_{K-1}\|^2\nonumber\\
			&\leq \mathbb{E}\|\tilde{x}-x^*\|^2-2\eta\sum_{k=0}^{K-1}\mathbb{E}g_k^T(x_{k}-x^*)+\eta^2\sum_{k=0}^{K-1}\mathbb{E}\|g_k\|^2\nonumber\\
			&\leq \mathbb{E}\|\tilde{x}-x^*\|^2+2\eta(\frac{8L\eta}{B}-1)\sum_{k=0}^K\mathbb{E}(P(x_{k+1})-P(x^*))+\frac{16L\eta^2}{B}(K+1)\mathbb{E}(P(\tilde{x})-P(x^*))\nonumber\\
			&\leq \frac{2}{\mu}\mathbb{E}(P(\tilde{x})-P(x^*))+2\eta(\frac{8L\eta}{B}-1)\sum_{k=0}^{K-1}\mathbb{E}(P(x_{k+1})-P(x^*))+\frac{16L\eta^2}{B}(K+1)\mathbb{E}(P(\tilde{x})-P(x^*))\nonumber\\\label{ineq4}
			\end{align}}
		where the last inequality follows by the strongly convexity assumption.
		
		By rearranging the Ineq. (\ref{ineq4}), we get:\\											
		{\small\begin{equation}
			2\eta(1-\frac{8L\eta}{B})\sum_{k=1}^{K}\mathbb{E}(P(x_{k})-P(x^*))\leq (\frac{2}{\mu}+\frac{16L\eta^2}{B})(P(\tilde{x})-P(x^*)) \label{ineq5}.
			\end{equation}}
		
		Dividing both sides of Ineq. (\ref{ineq5}) by $2\eta(1-\frac{8L\eta}{B})$, we obtain	
		{\small\begin{align*}
			&P(\tilde{x}_s)-P(x^*)\leq \left(\frac{B}{\eta\mu K(B-8\eta L)}+\frac{8\eta L}{(B-8\eta L)}\right)\mathbb{E}[P(\tilde{x}_{s-1})-P(x^*)].
			\end{align*}}

		\subsection{Proof of Theorem 4.3}	
		Let $x_{k+1}-x_{k}=-\eta g_k$, $v_{k}=\nabla f_{\mathcal{B}_k}(x_{k})-\nabla f_{\mathcal{B}_k}(\tilde{x})+\nabla F(\tilde{x})$, and $u_{k}=\nabla f_{\mathcal{B}_k}(\hat{x}_k)-\nabla f_{\mathcal{B}_k}(\tilde{x})+\nabla F(\tilde{x})$.
		Let  {\small$\bar{x}_{k+1}=argmin_{x\in R^d}\left\{\frac{1}{2}\|x-x_{k}-\eta u_k\|^2+\eta R(x)\right\}$}, and recall the following update rule for $x_k$:
		\begin{eqnarray*}
			&&x_{k+1,C_{j_k}}={prox_{\eta_kR_{j_k}}}\left\{x_{k,C_{j_k}}-\eta_k u_{k,C_{j_k}}\right\},\\
			&&x_{k+1,\setminus C_{j_k}}\leftarrow x_{k,\setminus C_{j_k}}.
		\end{eqnarray*}
		We take expectation with respect to $j_k$, and have,
		{\small\begin{align}
			&\mathbb{E}_{j_k}(x_{k+1}-x_{k})=\frac{1}{m}(\bar{x}_{k+1}-x_{k}) \label{eq1m}\\
			&\mathbb{E}_{j_k}\|x_{k+1}-x_{k}\|^2=\frac{1}{m}\|\bar{x}_{k+1}-x_{k}\|^2. \label{eq2m}\\\nonumber
			\end{align}}
		We have the following derivation for $P(x^*)$,
		{\small\begin{align}
			P(x^*)&=F(x^*)+R(x^*) \nonumber\\
			&\geq F(x_{k})+\nabla F(x_{k})^T(x^*-x_{k})+R(\bar{x}_{k+1})+\xi_{k+1}^T(x^*-\bar{x}_{k+1}) \nonumber\\\nonumber
			&=m\mathbb{E}_{j_k}P(x_{k+1})-(m-1)P(x_{k})+(\eta-\frac{T\eta^2}{2})m\mathbb{E}_{j_k}\|g_k\|^2+(v_k-\nabla F(x_{k}))^T(\bar{x}_{k+1}-x^*) \nonumber\\
			&\quad+m\mathbb{E}_{j_k}(g_k)^T(x^*-x_{k})+(\nabla f_{\mathcal{B}_k}(\hat{x}_k)-\nabla f_{\mathcal{B}_k}(x_{k}))^T(\bar{x}_{k+1}-x^*). \label{eq9m}\\\nonumber
			\end{align}}
		The first "$\geq$" holds, by the convexity of $F(x)$ and $R(x)$. The last "$=$" holds, by lemma B.1 in \cite{zhao2014accelerated}.
		
		Due to the delay, the Ineq.(\ref{eq9m}) has an extra term {\small$(\nabla f_{\mathcal{B}_k}(\hat{x}_k)-\nabla f_{\mathcal{B}_k}(x_{k}))^T(\bar{x}_{k+1}-x^*)$} compared to lemma B.1 in \cite{zhao2014accelerated}. Next we will show how to bound this term by using the separability of the coordinate blocks. Intuitively, each worker calculates a partial gradient at each iteration. When the  the number of block partitions $m$ is sufficient large, different workers select the same block with low probability.
		
		We decompose the term {\small$-(\nabla f_{\mathcal{B}_k}(\hat{x}_k)-\nabla f_{\mathcal{B}_k}(x_{k}))^T(\bar{x}_{k+1}-x^*)$} by using the partial smoothness assumption and the bounded and independent delay assumption as below:
		{\small\begin{align*}
			&\quad-(\nabla f_{\mathcal{B}_k}(\hat{x}_k)-\nabla f_{\mathcal{B}_k}(x_{k}))^T(\bar{x}_{k+1}-x^*) \nonumber\\
			&\leq\|\nabla f_{\mathcal{B}_k}(\hat{x}_k)-\nabla f_{\mathcal{B}_k}(x_{k})\|\|\bar{x}_{k+1}-x^*\|\nonumber\\
			&\leq T\|\hat{x}_k-x_k\|\|\bar{x}_{k+1}-x^*\|\nonumber\\
			&\leq T\sum_{h\subset J(k)}\|x_{h+1}-x_h\|\|\bar{x}_{k+1}-x^*\|\nonumber\\
			&\leq T\sum_{h=k-\tau}^{k-1}\|x_{h+1}-x_h\|\|\bar{x}_{k+1}-x^*\|\nonumber\\
			&=T\left\{\underbrace{\sum_{a=k-\tau}^{k-1}\|x_{a+1}-x_{a}\|\|\bar{x}_{k+1}-x_{k}\|}_{\textrm{(i)}}+\underbrace{\sum_{a=k-\tau}^{k-1}\sum_{b=a}^{k-1}\|x_{a+1}-x_{a}\|\|x_{b+1}-x_{b}\|}_{\textrm{(ii)}}+\underbrace{\sum_{a=k-\tau}^{k-1}\|x_{a+1}-x_{a}\|\|x_{a}-x^*\|}_{\textrm{(iii)}}\right\}.
			\end{align*}}
		For the term (\textrm{i}), we have
		{\small\begin{align}
			\sum_{a=k-\tau}^{k-1}\|x_{a+1}-x_a\|\|\bar{x}_{k+1}-x_{k}\|
			&\leq\sum_{a=k-\tau}^{k-1}\frac{1}{2}(\sqrt{m}\|x_{a+1}-x_{a}\|^2+\frac{1}{\sqrt{m}}\|\bar{x}_{k+1}-x_{k}\|^2)\nonumber\\
			&\leq \sum_{a=k-\tau}^{k-1}\frac{1}{2}(\sqrt{m}\|x_{a+1}-x_{a}\|^2+\sqrt{m}\mathbb{E}_{j_k}\|x_{k+1}-x_{k}\|^2)\nonumber\\
			&\leq \frac{\eta^2\sqrt{m}}{2}\sum_{a=k-\tau}^{k-1}\|g_a\|^2+\frac{\eta^2\tau\sqrt{m}}{2}\mathbb{E}_{j_k}\|g_k\|^2.\label{eq13}\\\nonumber
			\end{align}}
		The first "$\leq$" holds by the AM-GM inequality. The second "$\leq$" holds by Eqn (\ref{eq2m}).  \\
		
		For the term (\textrm{ii}), we have
		{\small\begin{align}
			\sum_{a=k-\tau}^{k-1}\sum_{b=a}^{k-1}\|x_{a+1}-x_{a}\|\|x_{b+1}-x_{b}\|
			&=\sum_{a=k-\tau}^{k-1}\eta^2\left(\|g_a\|^2+\sum_{b=a+1}^{k-1}\|g_a\|_{j_b}\|g_b\|\right)\nonumber\\
			&\leq \sum_{a=k-\tau}^{k-1}\eta^2\|g_a\|^2+\sum_{a=k-\tau}^{k-1}\eta^2\sum_{b=a+1}^{k-1}\left(\frac{\sqrt{m}\|g_a\|^2_{j_b}}{2}+\frac{\|g_b\|^2}{2\sqrt{m}}\right).\label{ineq7}
			\end{align}}
		It is clear that, $\mathbb{E}_{j_b}\|g_a\|_{j_b}^2=\frac{1}{m}\|g_a\|^2$ for $b>a$ since $j_b$ is independent to $g_a$.
		Therefore, the expectation of the Ineq.(\ref{ineq7}) inequality, we have the following derivation: 	
		{\small\begin{align}
			\mathbb{E}\sum_{a=k-\tau}^{k-1}\sum_{b=a}^{k-1}\|x_{a+1}-x_{a}\|\|x_{b+1}-x_{b}\|
			&\leq \mathbb{E}\sum_{a=k-\tau}^{k-1}\eta^2\|g_a\|^2+\sum_{a=k-\tau}^{k-1}\eta^2\sum_{b=a+1}^{k-1}\left(\frac{\sqrt{m}\mathbb{E}_{j_b}\|g_a\|^2_{j_b}}{2}+\frac{\|g_b\|^2}{2\sqrt{m}}\right)\nonumber\\\nonumber
			&=\mathbb{E}\sum_{a=k-\tau}^{k-1}\eta^2\|g_a\|^2+\sum_{a=k-\tau}^{k-1}\eta^2\sum_{b=a+1}^{k-1}\left(\frac{\|g_a\|^2}{2\sqrt{m}}+\frac{\|g_b\|^2}{2\sqrt{m}}\right)\nonumber\\
			&=\mathbb{E}\sum_{a=k-\tau}^{k-1}\eta^2(\frac{\tau}{2\sqrt{m}}+1)\|g_a\|^2 \label{eq14}\\\nonumber
			\end{align}}
		For the term (\textrm{iii}), we have
		{\small\begin{align}
			\sum_{a=k-\tau}^{k-1}\|x_{a+1}-x_{a}\|\|x_{a}-x^*\|
			&\leq\sum_{a=k-\tau}^{k-1}T\|x_{a+1}-x_{a}\|\|x_{a}-x^*\|_{j_a}\nonumber\\
			&\leq\sum_{a=k-\tau}^{k-1}\left(\frac{T\eta}{2\sqrt{m}}\|g_a\|^2+\frac{T\eta\sqrt{m}}{2}\|x_{a-1}-x^*\|_{j_a}^2\right).\label{ineq8}\\
			\end{align}}
		Taking expectations on both size of Ineq.(\ref{ineq8}), we can get:
		{\small\begin{eqnarray}
			\mathbb{E}\sum_{a=k-\tau}^{k-1}\|x_{a+1}-x_{a}\|\|x_{a}-x^*\|&\leq& \mathbb{E}\sum_{a=k-\tau}^{k-1}\left(\frac{T\eta}{2\sqrt{m}}\|g_a\|^2+\frac{T\eta\sqrt{m}}{2}\mathbb{E}_{j_a}\|x_{a}-x^*\|_{j_a}^2\right)\nonumber\\
			&=&\mathbb{E}\sum_{a=k-\tau}^{k-1}\left(\frac{T\eta}{2\sqrt{m}}\|g_a\|^2+\frac{T\eta}{2\sqrt{m}}\|x_{a}-x^*\|^2\right) \label{eq15}\\\nonumber
			\end{eqnarray}}
		Summing up Ineq. (\ref{eq13}),(\ref{eq14}) and (\ref{eq15}), we can get
		{\small\begin{align}
			&\quad-(\nabla f_{\mathcal{B}_k}(\hat{x}_k)-\nabla f_{\mathcal{B}_k}(x_{k}))^T(\bar{x}_{k+1}-x^*)\nonumber\\
			&\leq \frac{T\eta^2\tau\sqrt{m}}{2}\mathbb{E}\|g_k\|^2+\frac{T\eta}{2\sqrt{m}}\sum_{a=k-\tau}^{k-1}\mathbb{E}\|x_{a}-x^*\|^2+\sum_{a=k-\tau}^{k-1}\left(\left(\frac{T\eta}{2\sqrt{m}}+T\eta^2\left(\frac{\tau}{2\sqrt{m}}+1+\frac{\sqrt{m}}{2}\right)\right)\mathbb{E}\|g_a\|^2\right).\label{ineq14}\\\nonumber
			\end{align}}
		We have finished bounding the term $-(\nabla f_{\mathcal{B}_k}(\hat{x}_k)-\nabla f_{\mathcal{B}_k}(x_{k}))^T(\bar{x}_{k+1}-x^*)$.
		Taking expectation on both sides of Ineq. (\ref{eq9m}) and putting Ineq. (\ref{ineq14}) in Ineq. (\ref{eq9m}), we can get
		{\small\begin{align}
			&\quad-\mathbb{E}(g_k)^T(x_{k}-x^*)+(\eta-\frac{T\eta^2}{2}-\frac{T\eta^2\tau}{2\sqrt{m}})\mathbb{E}\|g_k\|^2 \nonumber\\
			&\leq -\left(\mathbb{E}P(x_{k+1})-P(x^*)\right)+\frac{(m-1)}{m}\mathbb{E}\left(P(x_{k})-P(x^*)\right)-\frac{1}{m}(v_k-\nabla F(x_{k}))^T(\bar{x}_{k+1}-x^*)\nonumber\\
			&\quad+\frac{T\eta}{2m^{\frac{3}{2}}}\sum_{a=k-\tau}^{k-1}\mathbb{E}\|x_{a}-x^*\|^2+\sum_{a=k-\tau}^{k-1}\frac{1}{m}\left(\frac{T\eta}{2\sqrt{m}}+T\eta^2\left(\frac{\tau}{2\sqrt{m}}+1+\frac{\sqrt{m}}{2}\right)\right)\mathbb{E}\|g_a\|^2.\label{eq16}\\\nonumber
			\end{align}}
		Summing up the Ineq. (\ref{eq16}) over $k=1,\cdots,K$, we have,
		{\small\begin{align*}
			&\quad\sum_{k=1}^{K}-\mathbb{E}(g_k)^T(x_{k}-x^*)+\left(\eta(1-\frac{T\tau}{2m^{\frac{3}{2}}})-T\eta^2(\frac{1}{2}+\frac{\tau}{2\sqrt{m}}+\frac{\tau}{m}+\frac{\tau^2}{2m^{\frac{3}{2}}})\right)\mathbb{E}\|g_k\|^2\\
			&\leq \sum_{k=1}^{K}-\left(\mathbb{E}P(x_{k+1})-P(x^*)\right)+\frac{(m-1)}{m}\mathbb{E}\left(P(x_{k})-P(x^*)\right)-\frac{1}{m}(v_k-\nabla F(x_{k}))^T(\bar{x}_{k+1}-x^*)\\
			&\quad+\frac{T\eta}{2m^{\frac{3}{2}}}\sum_{a=k-\tau}^{k-1}\mathbb{E}\|x_{a}-x^*\|^2\\
			&\leq \sum_{k=1}^{K}-\left(\mathbb{E}P(x_{k+1})-P(x^*)\right)+\frac{(m-1)}{m}\mathbb{E}\left(P(x_{k})-P(x^*)\right)-\frac{1}{m}(v_k-\nabla F(x_{k}))^T(\bar{x}_{k+1}-x^*)\\
			&\quad+\frac{T\eta}{\mu m^{\frac{3}{2}}}\sum_{a=k-\tau}^{k-1}\mathbb{E}(P(x_a)-P(x^*))\\
			&\leq \sum_{k=1}^{K}-\left(\mathbb{E}P(x_{k+1})-P(x^*)\right)+\sum_{k=1}^{K}(\frac{(m-1)}{m}+\frac{T\eta\tau}{\mu m^{\frac{3}{2}}})\mathbb{E}\left(P(x_{k})-P(x^*)\right)\\
			&\quad-\frac{1}{m}(v_k-\nabla F(x_{k}))^T(\bar{x}_{k+1}-x^*).
			\end{align*}}
		With the assumption $\eta<\frac{1}{T}\frac{m^{\frac{3}{2}}-T\tau}{m^{\frac{3}{2}}+3m\tau+\tau^2}$, we have $\eta(1-\frac{T\tau}{2m^{\frac{3}{2}}})-\eta^2T(\frac{1}{2}+\frac{(\tau-1)}{2\sqrt{m}}+\frac{\tau}{m}+\frac{\tau^2}{2m^{\frac{3}{2}}})>\frac{\eta}{2}$.
		Then, the above inequality can be reformulated as below,
		{\small\begin{align*}
			\sum_{k=1}^{K}-\mathbb{E}(g_k)^T(x_{k}-x^*)+\frac{\eta}{2}\mathbb{E}\|g_k\|^2&\leq \sum_{k=1}^{K}-\left(\mathbb{E}P(x_k)-P(x^*)\right)+\frac{4(K+1)\eta L}{mB}(P(\tilde{x})-P(x^*))\\
			&\quad+\sum_{k=1}^{K}(\frac{(m-1)}{m}+\frac{T\eta\tau}{\mu m^{\frac{3}{2}}}+\frac{4\eta L}{mB})\mathbb{E}\left(P(x_{k})-P(x^*)\right).\\
			\end{align*}}
		Therefore, we have the following upper bound for the sub-optimality,
		{\small\begin{align*}
			&\quad\mathbb{E}\|x_K-x^*\|^2\\
			&\leq \mathbb{E}\|\tilde{x}-x^*\|^2-2\eta\sum_{k=1}^{K}\left(\frac{1}{m}-\frac{T\eta\tau}{\mu m^{\frac{3}{2}}}-\frac{4\eta L}{mB}\right)\mathbb{E}\left(P(x_{k})-P(x^*)\right)+\frac{8K\eta^2 L}{mB}(P(\tilde{x})-P(x^*))\\
			&\leq -2\eta\sum_{k=1}^{K}(\frac{1}{m}-\frac{T\eta\tau}{\mu m^{\frac{3}{2}}}-\frac{4\eta L}{mB})\mathbb{E}\left(P(x_{k})-P(x^*)\right)+\left(\frac{2}{\mu}+\frac{8(K+1)\eta^2 L}{mB}\right)(P(\tilde{x})-P(x^*)).\\
			\end{align*}}
		By dividing both sides of the above inequality by $2\eta\sum_{k=1}^{K}(\frac{1}{m}-\frac{L\eta\tau}{\mu m^{\frac{3}{2}}}-\frac{4\eta L}{mB})K$ and choosing $B$ which satisfies $B>L/T$, we can obtain\\
		{\small\begin{align*}
			&P(\tilde{x}_s)-P(x^*)\leq \left(\frac{m}{\eta\mu K(1-\frac{T\eta\tau}{\mu m^{\frac{1}{2}}}-4\eta\alpha T)}+\frac{4\eta\alpha T(K+1)}{(1-\frac{T\eta\tau}{\mu m^{\frac{1}{2}}}-4\eta\alpha T)K}\right)\mathbb{E}[P(\tilde{x}_{s-1})-P(x^*)].
			\end{align*}}
		\subsection{Additional Experiments}

		We conduct experiments for comparing Async-ProxSVRG and Async-ProxSVRCD with other asynchronous proximal algorithms: Async-ProxSGD and Async-ProxSVRCD.
		For all the experiments, we set the number of local workers $P=10$. The parameter settings for Async-ProxSVRG and Async-ProxSCRCD are the same as the settings in section 5 in paper "Asynchronous Stochastic Proximal Optimization Algorithms with Variance Reduction". We use a decreasing step size for ProxSGD with $\eta=\eta_0\sqrt{\frac{\sigma_0}{t+\sigma_0}}$\cite{reddi2015variance}, where constant $\eta_0$ and $\sigma_0$ specify the scale and speed of decay.  Since $L$ has an upper bound of $0.25$. We set the step size for ProxSCD with $\eta=\frac{1}{T}\approx\gamma\sqrt{d}$ and we choose $\gamma=0.4$.
		
\begin{figure}[t]
		\centering
		\subfigure[]{
			\label{figa1}
			\includegraphics[width=1.6in]{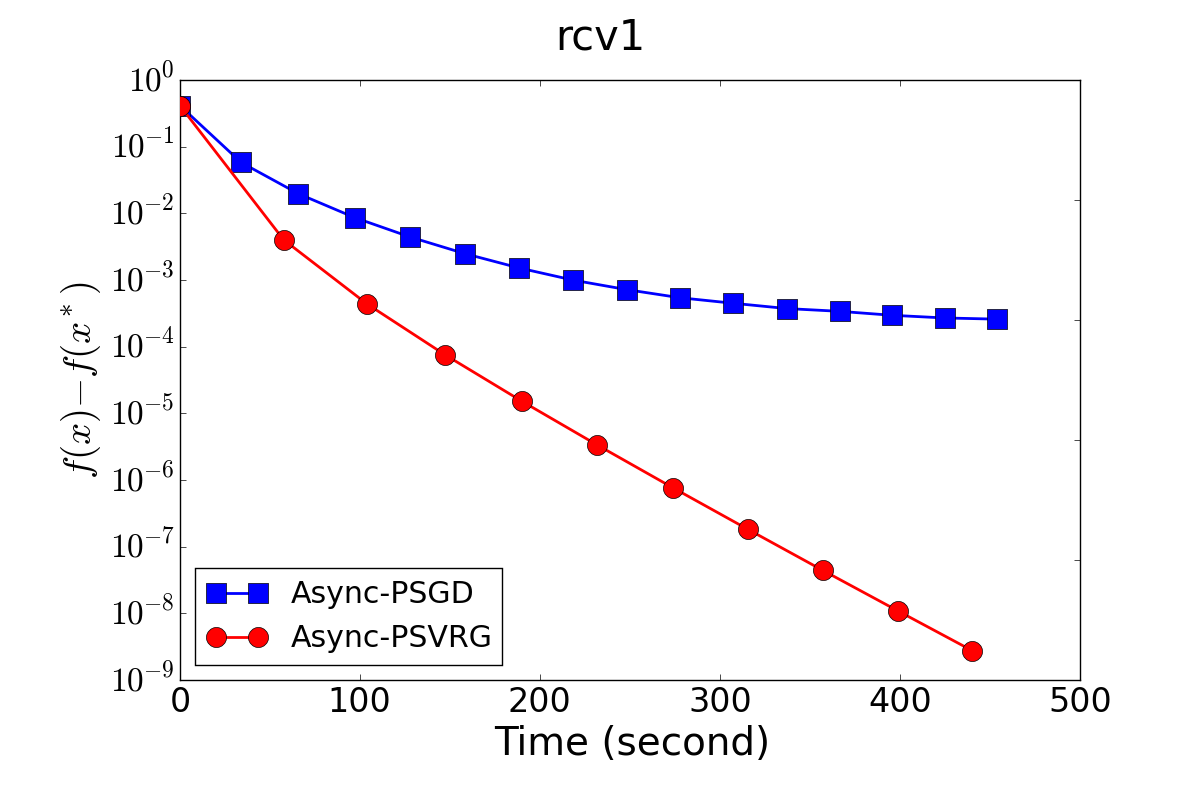}}
		\subfigure[]{
			\label{figa2}
			\includegraphics[width=1.6in]{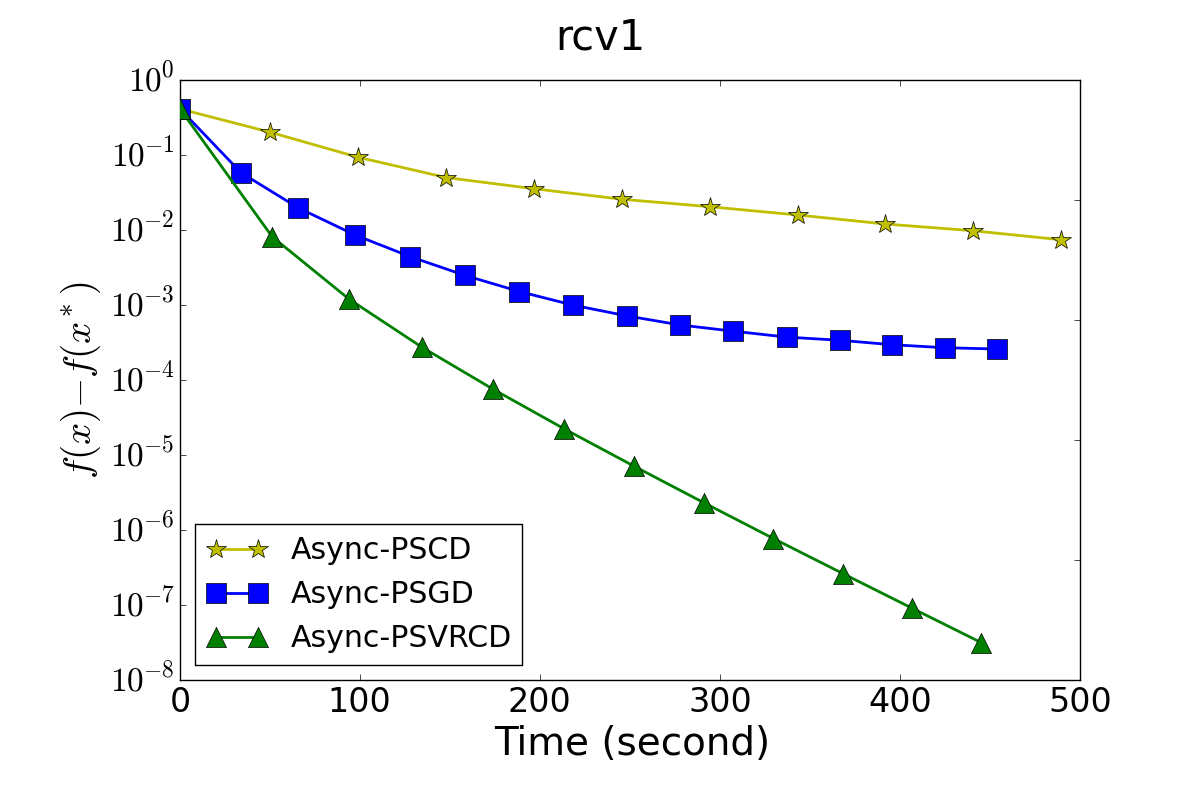}}
		\subfigure[]{
			\label{figb1}
			\includegraphics[width=1.6in]{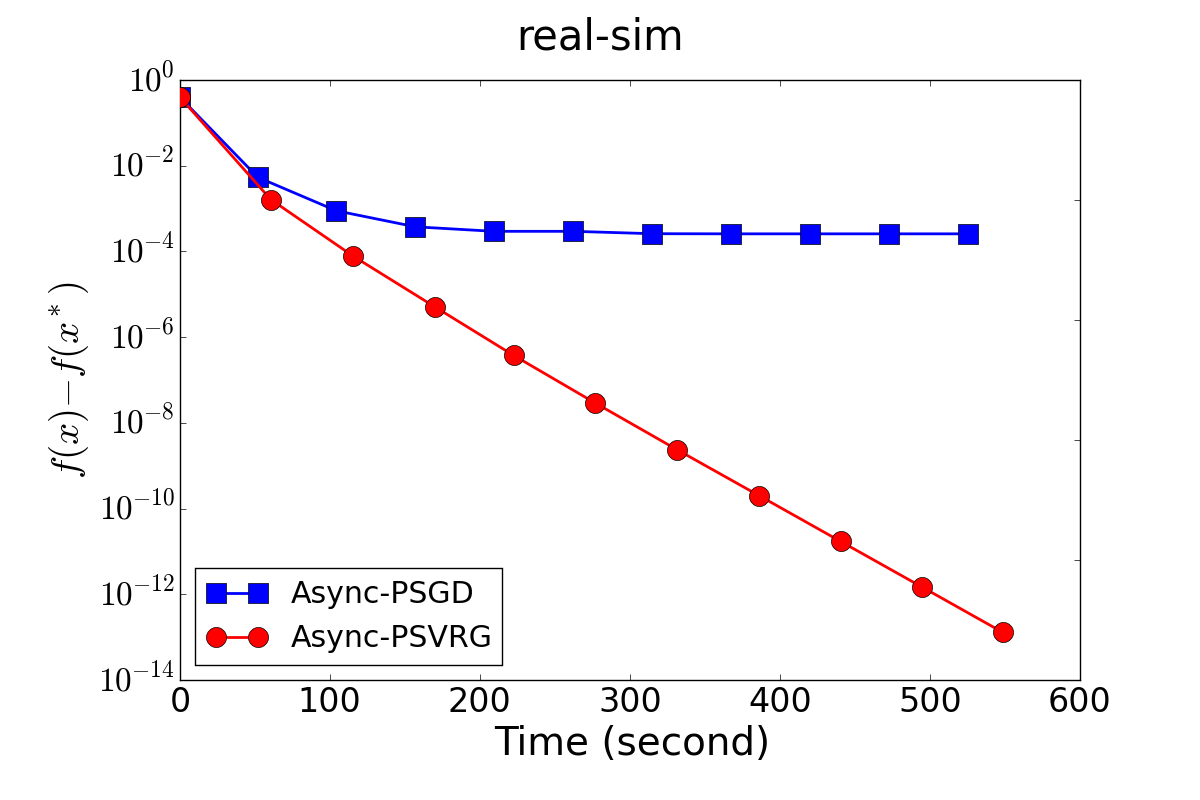}}
		\subfigure[]{
			\label{figb2}
			\includegraphics[width=1.6in]{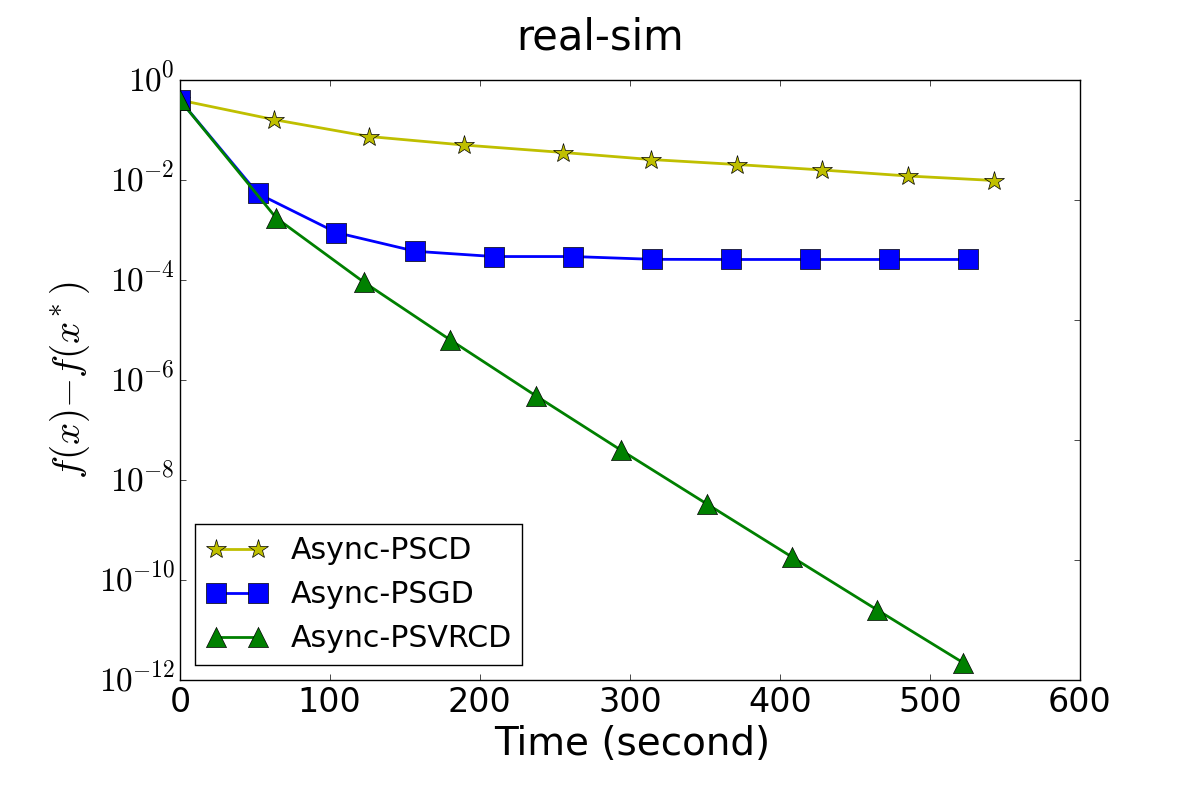}}
		\subfigure[]{
			\label{figc1}
			\includegraphics[width=1.6in]{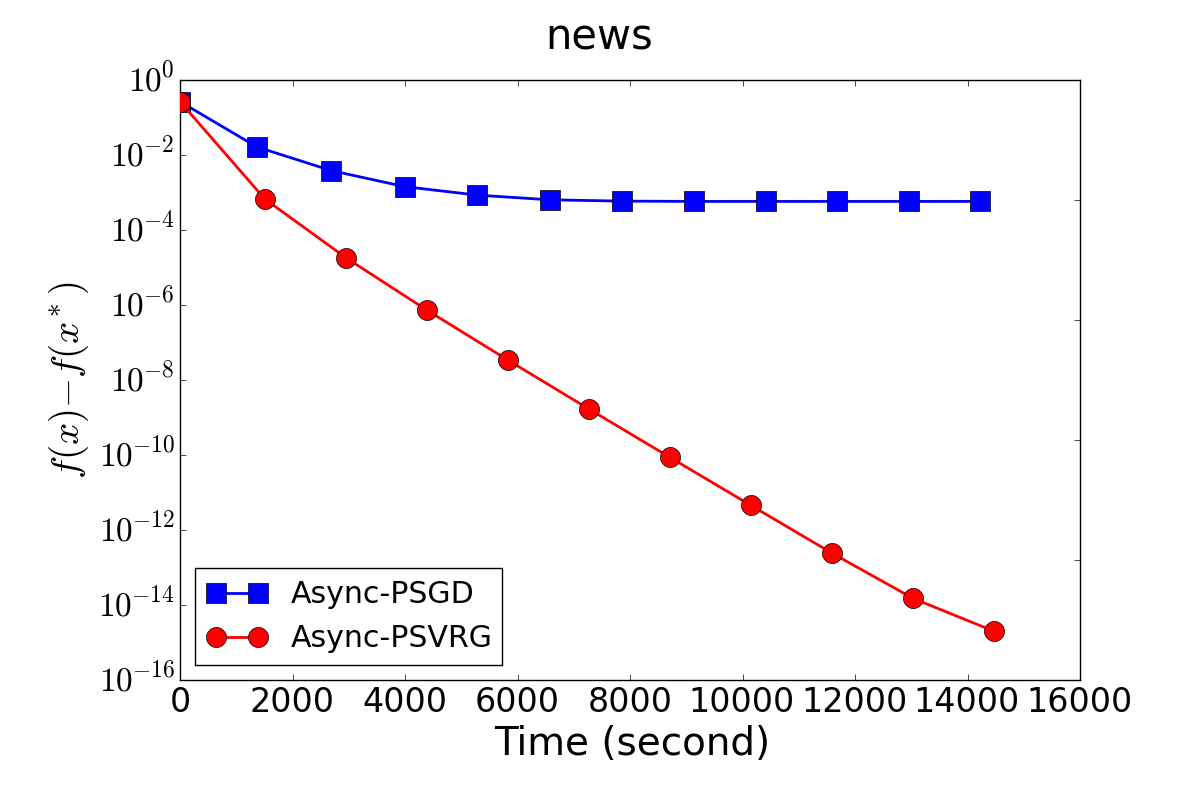}}
		\subfigure[]{
			\label{figc2}
			\includegraphics[width=1.6in]{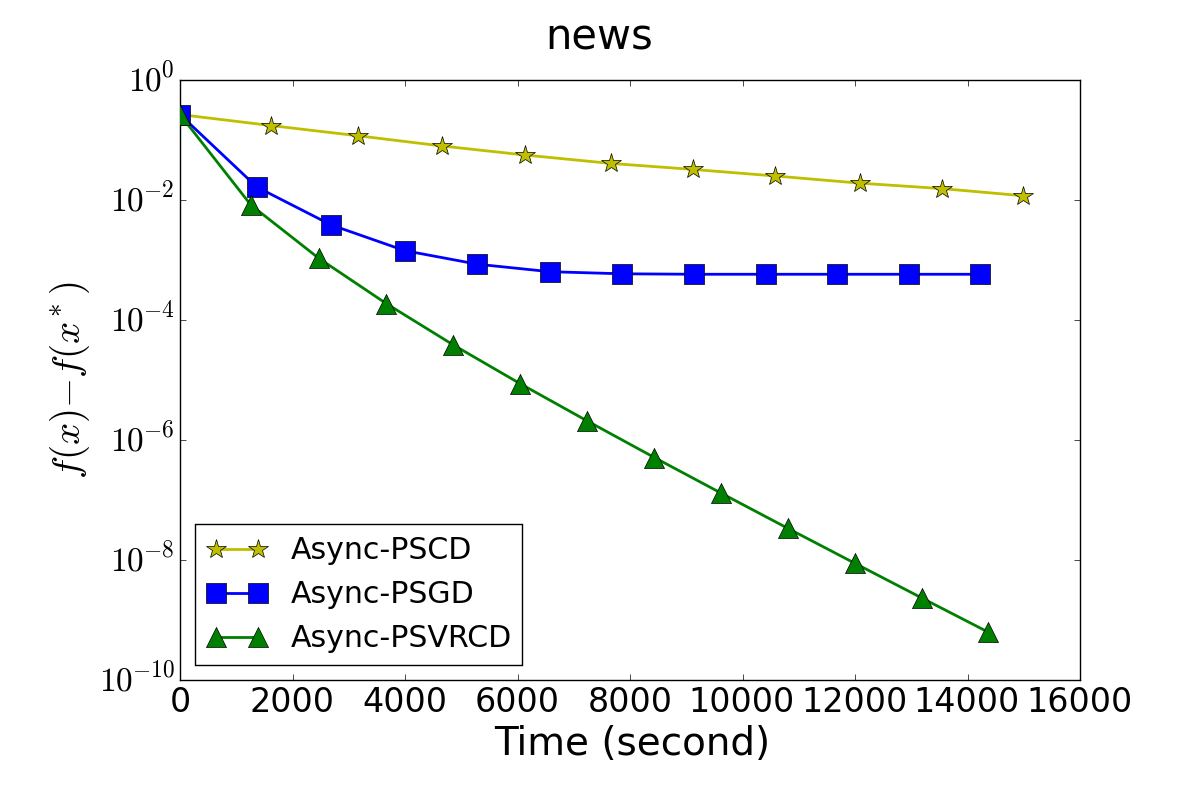}}
		\caption{(\ref{figa1}),(\ref{figb1}) and (\ref{figc1}) are comparison for Async-ProxSVRG with Async-SGD; (\ref{figa2}), (\ref{figb2}) and (\ref{figc2}) are comparison for Async-ProxSVRCD with Async-SGD and Async-SCD}
		\label{figure 1}
	\end{figure}

		The results are showed in Figure \ref{figure 1}. Figure \ref{figa1}, \ref{figb1} and \ref{figc1} show the comparison between Async-ProxSVRG and Async-ProxSGD on different data sets. The results show that Async-ProxSVRG outperforms Async-ProxSGD on all the three data sets. Figure \ref{figa2},\ref{figb2} and \ref{figc2} show the comparison between Async-ProxSVRCD and Async-ProxSGD and Async-ProxSCD.The results show that Async-ProxSVRG outperforms other algorithms on all the three data sets. It means that our proposed algorithms are efficient.
			
	\end{document}